\documentclass[iicol]{sn-jnl}

\usepackage[caption=false,font=normalsize,labelfont=sf,textfont=sf]{subfig}
\usepackage{array}
\usepackage{tabularx}
\usepackage{textcomp}
\usepackage{stfloats}
\usepackage{url}
\usepackage{verbatim}
\usepackage{graphicx}
\usepackage{multirow}
\usepackage{amsmath,amssymb,amsfonts,bm}
\usepackage[lighttt]{lmodern}
\usepackage{hyperref}
\usepackage{amsthm}%
\usepackage{mathrsfs}%
\usepackage[title]{appendix}%
\usepackage{xcolor}%
\usepackage{textcomp}%
\usepackage{manyfoot}%
\usepackage{booktabs}%
\usepackage{algorithm}%
\usepackage{algorithmicx}%
\usepackage{algpseudocode}%
\usepackage{listings}%
\usepackage{dsfont}
\usepackage{rotating}
\usepackage{enumitem}
\usepackage{tabularx}
\usepackage{multirow}
\usepackage{multicol}
\usepackage{threeparttable}
\usepackage{pifont}  
\usepackage{blindtext}
\usepackage{float}
\usepackage{tcolorbox}
\newcommand{\xmark}{\ding{55}}

\theoremstyle{thmstyleone}%
\theoremstyle{thmstyletwo}%

\theoremstyle{thmstylethree}%

\begin{document}

\title[An Adaptive Supervised Contrastive Learning Framework for Implicit Sexism Detection in Digital Social Networks]{An Adaptive Supervised Contrastive Learning Framework for Implicit Sexism Detection in Digital Social Networks}


\author[1]{\fnm{Mohammad Zia Ur} \sur{Rehman}}\email{phd2101201005@iiti.ac.in}\equalcont{These authors contributed equally to this work.}

\author[2]{\fnm{Aditya} \sur{Shah}}\email{c2k21106769@ms.pict.edu}
\equalcont{These authors contributed equally to this work.}

\author[1]{\fnm{Nagendra} \sur{Kumar}}\email{nagendra@iiti.ac.in}

\affil[1]{\orgdiv{Department of Computer Science and Engineering}, \orgname{Indian Institute of Technology Indore}, \orgaddress{ \city{Indore}, \postcode{453552}, \state{Madhya Pradesh}, \country{India}}}

\affil[2]{\orgdiv{Department of Computer Engineering}, \orgname{Pune Institute of Computer Technology}, \orgaddress{ \city{Pune}, \postcode{411043}, \state{Maharashtra}, \country{India}}}


\abstract{

The global reach of social media has amplified the spread of hateful content, including implicit sexism, which is often overlooked by conventional detection methods. In this work, we introduce an Adaptive Supervised Contrastive lEarning framework for implicit sexism detectioN (ASCEND). A key innovation of our method is the incorporation of threshold-based contrastive learning: by computing cosine similarities between embeddings, we selectively treat only those sample pairs as positive if their similarity exceeds a learnable threshold. This mechanism refines the embedding space by robustly pulling together representations of semantically similar texts while pushing apart dissimilar ones, thus reducing false positives and negatives. The final classification is achieved by jointly optimizing a contrastive loss with a cross-entropy loss. Textual features are enhanced through a word-level attention module. Additionally, we employ sentiment, emotion, and toxicity features. Evaluations on the EXIST2021 and MLSC datasets demonstrate that ASCEND significantly outperforms existing methods, with average Macro F1 improvements of 9.86\%, 29.63\%, and 32.51\% across multiple tasks, highlighting its efficacy in capturing the subtle cues of implicit sexist language.\\ 
\textbf{[Content Warning]}:\textbf{This paper contains examples of sexist language for illustration. These examples may be offensive to some readers. We do not promote the views presented in these examples.}\\}




\maketitle

\section{Introduction}
\label{sec:sa_int}
Social media has changed the way people communicate with each other on a daily basis, allowing people to stay in touch with each 
other. According to Statista \cite{Petrosyan2024}, 62.2\% of the world’s population uses social media. It has allowed people to exchange information and opinions at an enormous scale. However, this has also led to an increase in cyberbullying. People use derogatory terms and indulge in the use of hate speech and other harmful behaviour. While social media platforms implement systems to curb such content, reports of hate speech and sexist content have only increased through the years \cite{rehman2023user}. Most social media platforms flag such content by using machine learning systems, and then rely on humans to manually analyze the content to determine the appropriate action. Platforms such as Twitter (X) and Instagram, generate 500 million tweets and 1.3 billion posts respectively, on a daily basis. Consequently, this manual review process is extremely time-consuming. By the time the content is addressed, the damage may already have been done. This creates a harmful environment online, taking a toll on one’s emotional and mental well-being. Preventive systems, capable of directly analyzing the content are the need of the present.
The large volume of content generated and the intricate structure of languages make recognizing such content a particularly difficult task.

To ensure a healthy and safe experience while using social media, it is necessary to combat this explosion in sexist content.  Addressing this issue requires a robust system capable of detecting sexism across different languages and in a variety of contexts, as only 52.1\% of the web content is in English, while the remaining is split between languages such as Spanish (5.5\%),  German (4.8\%), Russian (4.5\%), Japanese (4.3\%), French (4.3\%) and more. To replicate such environments, we use datasets consisting of tweets \cite{RodrguezSanchez2021OverviewOE} and online accounts of sexism \cite{parikh2019multi}, carefully annotated by experts. These datasets contain exhaustive examples of implicit forms of sexist content. This careful annotation by experts ensures that our system has seen high-quality relevant data, improving its ability to detect sexism.

Sexist content is often context-dependent and has language-specific features and subtle structures. Previous research \cite{dutta2021efficient} \cite{samghabadi2020aggression} has majorly focused on detecting explicit forms of sexism. Research on fine-grained, implicit forms of sexism has a lot of potential. Implicit forms of sexism can be hard to detect because they contain subtle cues. A comment that seems harmless might still have hints of sexism, making it challenging to accurately classify the content. Moreover, what is considered sexist in one language may not be considered so in another language. Explicit sexism, due to its inherently extreme nature, is often met with immediate backlash. However, implicit sexism can go unchallenged, spreading stereotypes and biases.

\begin{table*}[h!]
 \centering
\begin{tabular}{|l|l|}
\hline
\hline
\multicolumn{1}{|l|}{\textbf{Type of Speech}}                   & \textbf{Example}               \\     
\hline \hline
(i) Sexist Speech & ``call me sexist but I've never seen a girl eat a whole \\ & large dominos pizza by herself'' \\
\hline
(ii) Non-Sexist Hate-Speech & ``People will buy a ring light as if it changes their \\ & appearance. You're still ugly it's just easier for us to see now'' \\
\hline
(iii) Non-Hate Speech & ``the people deserve to know why they are being denied \\ & \$2,000 / month survival checks. COVID-19 relief should \\ & not be debated behind closed doors'' \\
\hline
(iv) Sexist Speech & ``Well hello isn’t kathys skirt very short'' \\
\hline
\end{tabular}
\vspace{0.5em}
\caption{Types of Speech and Examples}
\label{table:speech_types_example}
\end{table*}

Table-~\ref{table:speech_types_example} demonstrates examples of sexist and non-sexist speech. The first and fourth samples are examples of \textit{Sexist Speech} as the former insinuates that women have small appetites while the latter is objectifying the person. On the other hand, the second sample is classified as \textit{Non-Sexist Hate-Speech} since it tries to belittle and humiliate a person, albeit avoiding derogatory terms. As the third sample does not make use of any derogatory language, it is classified as \textit{Non-Hate Speech}

To tackle this problem, we propose a method \textit{Adaptive Supervised Contrastive lEarning framework for implicit sexism detectioN in Digital social networks (ASCEND)}, designed to analyze the sexist content generated on social media platforms. This method leverages exhaustive context-sensitive examples ranging from explicit to implicit forms of sexism.
We develop a deep learning architecture, capable of detecting and categorizing implicit sexist content. Our architecture incorporates transformer models such as RoBERTa \cite{liu2019robertarobustlyoptimizedbert} in a contrastive learning based paradigm. Transformer based models understand the context and the nuances of language \cite{rehman2025hierarchical}. 
To overcome the limitations of previous approaches in implicit environments, we incorporate an adaptive supervised contrastive learning framework to help the system understand and detect implicit forms of sexism, enabling the system to distinguish between sexist and non-sexist content, even when the differences are subtle.

The following points summarize our key contributions:
\begin{enumerate}
\item{This paper presents a novel supervised-contrastive learning framework, ASCEND, to detect and categorize implicit sexist content. We propose an architecture capable of identifying, categorizing, and tagging implicit sexist content across various classification environments.}
\item{Leveraging contrastive learning makes our model understand the nuances and subtle semantics of sexist content, by creating an embedding space where similar features are located close to each other and far away from dissimilar features.}
\item{Our proposed method also employs word-level attention to accurately capture the interaction between features, providing additional context during classification. This mechanism allows the system to focus on the most relevant part of the content, helping to capture the context with higher precision.}
\item{Furthermore, we enhance the contrastive learning approach by introducing thresholding to create a more robust embedding space. We only consider features as similar if they express high cosine similarity, even if both features share the same label, reducing the occurrence of false-positives and false-negatives.}
\item{Experimentation across two datasets shows that ASCEND  outperforms the current state-of-the-art systems across a variety of implicit classification domains, namely sexism detection, sexism categorization, and multi-label sexism classification.}
\end{enumerate}

The paper is organized as follows:
Section 1 provides an introduction to the topic, the motivation for this research, and why it is necessary. Section 2 reviews the related works on this topic, highlighting their strengths and weaknesses. Section 3 highlights our proposed method, while Section 4 provides quantitative and qualitative results of this system. Finally, we present our conclusion in Section 5.

\section{Related Work}
\label{sec:sa_rw}
In this section, we review previous literature and describe their methodology in an attempt to identify their pros and cons.
\subsection{Machine Learning and Deep Learning Methods}
Waseem and Hovy \cite{waseem2016hateful} used character n-grams to train a logistic regression classifier on a custom dataset, detailing its construction and annotation.
Alfina et al.  \cite{alfina2017hate} compared the performance of traditional machine learning approaches, namely Naive Bayes, Support Vector Machines, Bayesian Logistic Regression, and Random Forest Classifiers for the task of hate speech detection in low-resource languages such as the Indonesian language.
However, these traditional methods require manual feature engineering and struggle to generalize to new languages, leading to lower accuracy in multilingual applications. 
Badjatiya et al.  \cite{badjatiya2017deep} experimented with different deep learning architectures such as deep neural networks, CNN and LSTM. These architectures outperformed traditional machine learning-based approaches by automatically learning features from the data, reducing the need for manual feature engineering. Zhang et al.  \cite{zhang2018detecting} used a combination of CNN and Gated Recurrent Unit (GRU) \cite{cho2014learning} for detecting hate speech. 

Nevertheless, deep learning-based approaches fail to capture the subtleties and the finer semantics of the language because sexist content in social networks is often nuanced and difficult to capture.
After the introductions of transformers \cite{vaswani2017attention} and BERT \cite{devlin2018bert}, transfer learning techniques have been used to harness the power of large pre-trained transformer models. 
 Researchers have been able to adapt these models to this specific task by fine-tuning models such as BERT. \ Parikh, et al.\cite{parikh2019multi} demonstrated that a BERT model combined with hierarchical recurrent and convolutional operations outperformed deep learning approaches. 
While transformers demonstrated strong performance across multiple languages and captured context effectively, they did not maintain this performance on tasks requiring fine-grained analysis and categorization. A few approaches have also been proposed for multimodal sexism detection\cite{rehman2025context}.

\subsection{Contrastive Learning methods}
Recently, researchers have also incorporated contrastive learning for the task of detecting hateful content. Contrastive learning creates a feature space where similar features are grouped together and dissimilar features are distant, making it easier for the model to distinguish between these features and use them during downstream analysis \cite{dar2024contrastive, raghaw2024explainable}. It has primarily been applied to the domain of computer vision \cite{chen2020simple},\cite{khosla2020supervised}, 
 \cite{diba2021vi2clr}, \cite{grill2020bootstrap}, but recently researchers have adapted this technique to textual data. 

Gao et al.  \cite{gao2021simcse} used contrastive learning to enhance sentence representations in textual data. Their experiments also indicated that contrastive learning makes pre-trained word embeddings more uniform and leads to better alignment of positive pairs when labeled data is available. Angel, et al.  \cite{angel2023multilingual} combined contrastive learning with regression by integrating the annotator information to predict the number of annotators that classify an example as positive or negative. This was used as an intermediate step in the conventional transformer fine tuning approach.
Lu et al.  \cite{lu2023hate} used a dual contrastive learning framework for hate speech detection. They also integrated focal loss into this framework of self-supervised and supervised learning to compensate for the data imbalance, achieving state-of-the-art results. 
Despite this, dual contrastive learning involves the training of multiple loss functions, which can be complex to implement and challenging to optimize. Training multiple objectives simultaneously requires careful consideration.

\section{Methodology}
This section delineates the architecture of our proposed approach. It covers the tasks of data pre-processing, feature extraction, and the incorporation of contrastive learning.

\begin{figure*}[h]
    \centering
    \includegraphics[width=16.5cm, height=11.1cm]{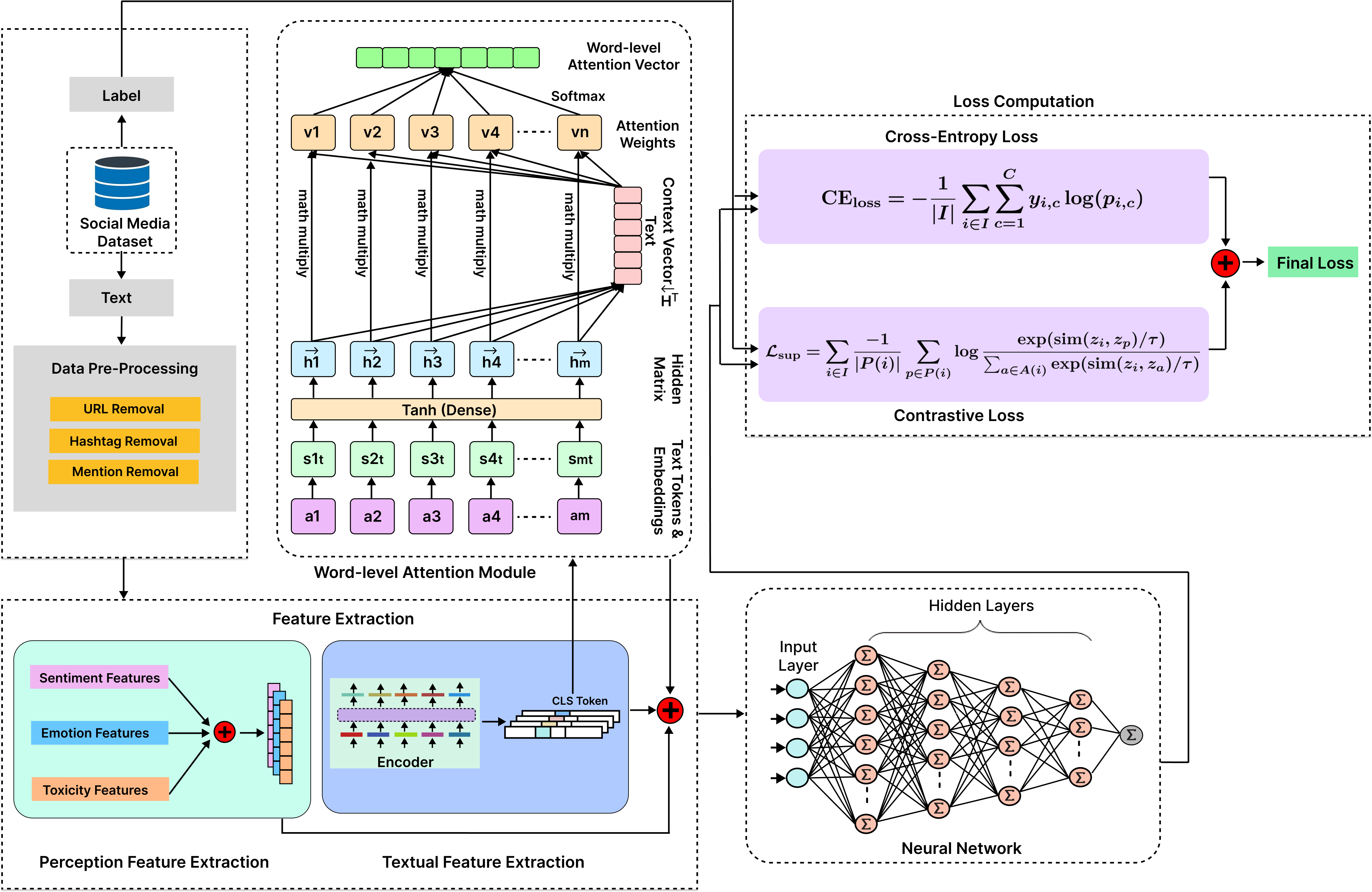}
    \caption{System Architecture}
    \label{fig:method}
\end{figure*}

\subsection {Data Pre-processing}
The data present in social media platforms often contains elements such as Uniform Resource Locators (URLs), hashtags, and mentions. These elements do not carry any significant meaning or context. They introduce noise into the data, affecting the performance of the learning algorithm.  Thus, we remove the
URLs, mentions, and hashtags from the data using regular expressions. Additionally, extra white spaces and special characters are also removed \cite{rehman2023kisanqrs}.

\subsection {Feature Extraction} 

\subsubsection{Textual Feature Extraction} 
To extract the features from the input data, we employ the transformer model RoBERTa \cite{liu2019robertarobustlyoptimizedbert}. It generates embeddings that are context-sensitive, capturing the meaning of a token in its entirety.
RoBERTa is made up of an encoder and a decoder. The encoder provides a continuous representation of the input sequence. The decoder then uses this representation for downstream tasks such as classification or summarization. We use the encoder as our feature extractor.
Given an input sequence W, consisting of x words, it is first tokenized to break the sequence down into tokens. The RoBERTa tokenizer produces a list of integers B, based on the model’s predefined vocabulary. Two unique tokens are added to each input sequence: a classification token [CLS] is added at the start of the input sequence, and a separator token [SEP] is added at the end of the sequence \cite{bansal2024hybrid}. 
\begin{equation}
B = \text{RoBERTa\_Tokenizer}(W)
\end{equation}
\begin{equation}
B = \{b_x\}_{x=1}^X
\end{equation}

Here, the sequence is limited to a maximum number of tokens X. If the length of the sequence is more than X, it is truncated. Alternatively, if the length of the sequence is less than X, it is padded to reach this length.

The sequence of tokens is then passed to the embedding layer, which converts the tokens into dense vectors. The embedding layer uses pre-trained word embeddings from the RoBERTa model. The embedded tokens are then passed through stacked transformer encoder layers. The RoBERTa-large model consists of 24 layers, each with 16 attention heads. Each attention head focuses on different parts of the input. Each layer uses self-attention to capture the context and interplay between the tokens in the sequence. We extract the hidden states from the last layer of the encoder. These hidden states are a 1024-dimensional vector. Along with the hidden states, we also extract the representation of the [CLS] token, which captures the overall meaning of the entire input sequence.

\begin{equation}
L = \text{RoBERTa}(B) 
\end{equation}
\begin{equation}
L = \{e_x\}_{x=1}^X 
\end{equation}
where  $L$ is the last-hidden-state representation and  $e_x$ is the embedding generated by RoBERTa.

\subsubsection{Perception Feature Extraction Extraction}
To enhance the capabilities of our system, we make use of sentiment, toxicity, and emotional features. 

\begin{enumerate}
    \item \textit{Sentiment Features:} 
To extract the sentiment features from a textual sequence, we employ the VADER lexicon \cite{hutto2014vader}. Vader is a rule-based tool that computes and normalizes the ratings between -1 (extremely negative) and +1 (extremely positive). A rating of 0 reflects that the statement is neutral in nature. For instance, consider the statement \textit{“I want a doll cuz i hate people, not women, I don't feel good talking with people i don't know, so my chance of getting a GF is zero, mostly because i don't want to be hurt like my first and last time in a relationship.”}
\newline  Vader generates the following Sentiment Features - \\ \textit{{`neg': 0.192, `neu': 0.699, `pos': 0.109, `compound': -0.6858}}. Since the above text is strongly negative in nature, the \textit{compound} (overall) nature of the text is -0.6858.

\item \textit{Emotion Features:} 
We utilize the NRCLex lexicons \cite{mohammad2013crowdsourcing} to distill the emotional features from the input sequence. It measures the \textit{fear, anger, anticipation, trust, surprise, positive, negative, sadness, disgust,} and \textit{joy} to analyze the emotions. Each category is assigned a score between 0 and 1, reflecting its severity. Consider the sentence \textit{“Why hate on a beautiful woman making her money? I rather learn from them. That’s the difference between you and me. I admire women while you hate and try to shade them. Do better.”}
\newline Emotion Features generated are - \\  
\textit{{`fear': 0.0333, `anger': 0.0333, `trust': 0.1333, `surprise': 0.1, `positive': 0.2333, `negative': 0.1333, `sadness': 0.0333, `disgust': 0.0333, `joy': 0.1333, `anticipation': 0.1333}}

In this example, the positive emotion is assigned the highest score since the sentence evokes an overall positive feeling, despite its varied nature.

\item \textit{Toxicity Features:} 
The ToxicBert model on huggingface \footnote{https://huggingface.co/unitary/toxic-bert} was used to obtain the toxicity features of a sentence. It measures the following categories: toxic, insult, obscene, identity hate, threat and severe toxic. Each category is then assigned a score between 0 and 1 depending on its intensity. For instance, consider the sample \textit{“Im a Sagittarius so I hate Sagittarius women”}
\newline Following features are generated - \\ 
\textit{{`toxic': 0.9567, `severe\_toxic': 0.03366, `obscene': 0.1949, `threat': 0.0165, `insult': 0.2927, `identity\_hate': 0.5879}}

The statement is extremely venomous, necessitating a very high toxic score

\end{enumerate}
These features are concatenated to the vector representation of the input text, providing additional information to the system about the nature of the input. 

\subsection {Contrastive Learning} 
Supervised Contrastive Learning, Khosla et al. (2020) \cite{khosla2020supervised}, incorporates information about the labels to create an embedding space, by pulling together features belonging to the same class and pushing away the features belonging to the other classes. The class labels are used to determine which pairs of features are positive and which are negative. A contrastive learning loss function is used to guide the learning process.

\subsubsection{Adaptive Contrastive Learning} 
For classification tasks, the [CLS] features within the embedding space are used to optimize the contrastive loss. We incorporate supervised contrastive learning to deal with the implicit 
nature and nuanced structure of our samples.

A binary matrix of size $\text{batch\_size} \times \text{batch\_size}$ is created. If samples $S_i$ and $S_j$ in a batch share the same label, $\text{matrix}[i][j]$ is set to 1. Consequently, if $S_i$ and $S_j$ possess different labels, $\text{matrix}[i][j]$ is set to 0. This matrix metadata is used to compute the contrastive loss.

Let $I$ denote the set of all indices in the batch. $P(i)$ represents the set of indices of all positive samples (samples from the same class as $i$) in the batch, excluding $i$, and $A(i)$ denote the set of indices of all samples in the batch, excluding $i$. $z_i$ represents the representation (embedding) of the $i$-th sample. We compute the similarity score between the representations $z_i$ and $z_j$ using cosine similarity. $\tau$ is the temperature parameter that scales the similarity score.
The contrastive loss is computed as:
\begin{equation}
\mathcal{L}_{\text{sup}} = \sum_{i \in I} \frac{-1}{|P(i)|} \sum_{p \in P(i)} \log \frac{\exp(\text{sim}(z_i, z_p) / \tau)}{\sum_{a \in A(i)} \exp(\text{sim}(z_i, z_a) / \tau)}
\end{equation}

Given representations $z_i$ and $z_j$, we calculate their cosine similarity as:

\begin{equation}
\text{cosine similarity}(z_i, z_j) = \frac{z_i \cdot z_j}{|z_i| |z_j|}
\end{equation}

$\mathbf{z}_i \cdot \mathbf{z}_j$ is the dot product of $\mathbf{z}_i$ and $\mathbf{z}_j$, $|\mathbf{z}_i|$ and $|\mathbf{z}_j|$ represent the magnitude of $\mathbf{z}_i$ and $\mathbf{z}_j$ respectively.

We improve the robustness of our system through thresholding. Features are considered similar only if they express high cosine similarity, even if both features share the same label. This reduces the occurrence of false-positives and false-negatives.

We populate the matrix by calculating the cosine similarity for each sample pair in the batch. Samples $S_i$ and $S_j$ are considered positive only if $\text{matrix}[i][j]$ is 1 and their cosine similarity is above a threshold. This threshold is set as a parameter, which is learned by the model during training.

\subsubsection{Word Level Attention}
Transformers utilize self-attention to understand the relationship between the words in an input sequence. However, for the task of detecting sexism, certain terms in the input sequence may be more significant and important than others. To capture this importance, we employ word-level attention \cite{yang2016hierarchical}. Word-level attention assigns importance to each word in a sequence, allowing the system to weigh the importance of each word when generating the output. This importance is also context-sensitive due to the inherent nature of the task. 
The input tokens are first fed to the encoder to generate dense vectors. Word-level attention operates on top of these embeddings to generate the weights for each token. 

Furthermore, to enhance the quality of the classifier parameters, we also employ the use of the word-level attention (WLA) vector on the last hidden state representation of a sample. This sentence representation incorporates both global and local context.

To calculate the word level attention vector $s_i$ for i, we use its hidden state $u_it$ and a word level context vector $u_w$ to get a normalized importance vector $it$. The sentence representation $s_i$ is calculated using the weighted sum of its words, based on their weights, as follows:

\begin{equation}
\alpha_{it} = \frac{\exp(u_{it}^\top u_w)}{\sum_t \exp(u_{it}^\top u_w)}
\end{equation}
\begin{equation}
\text{wla\_feats} = \sum_t \alpha_{it}h_{it}
\end{equation}

\begin{equation}
\text{wla\_feats} = \text{Word Level Attention}(\text{Last Hidden State})
\end{equation}

Following this,  we concatenate the word-level attention weights with the [CLS] token representation.

We then calculate the contrastive loss using the sentence representation. This adaptive contrastive learning approach enhances the model's ability to distinguish between similar and dissimilar features, even when the differences are subtle, thereby improving the detection and categorization of implicit sexist content.

\subsection{Classification}
For an input sequence $i$, with $t$ tokens and ground truth label $y$, we extract its hidden state representation $u_{it}$. The model makes a prediction $\hat{y}$ for the sample.

We optimize the model’s predictions by aligning them with the ground truth values using the cross-entropy loss. It minimizes the differences between the predictions and the actual labels.

\begin{equation}
\text{CE}_{\text{loss}} = \text{CrossEntropy\_Loss}(y, \hat{y})
\end{equation}

Let us assume there are $|I|$ samples in a batch and there are $C$ classes. $p_{i,c}$ is the predicted probability that sample $i$ belongs to class $c$, and $y_{i,c}$ is the ground-truth label of sample $i$.

\begin{equation}
\text{CE}_{\text{loss}} = -\frac{1}{|I|} \sum_{i \in I} \sum_{c=1}^C y_{i,c} \log(p_{i,c})
\end{equation}

The final loss is calculated as a combination of the contrastive loss and the cross-entropy loss:
\begin{equation}
\text{Loss} = \text{CL}_{\text{loss}} + \text{CE}_{\text{loss}}
\end{equation}

In the sexism\_detection function of Algorithm \ref{algo:sexism_detection}, line 2 tokenizes the input sequence, to prepare it for the RoBERTa transformer model. In line 3 to line 6, the last-hidden-state representation and [CLS] token representation are extracted. Line 7 and line 8 incorporate word-level-attention representation of the [CLS] token. Lines 9 through 11 extract the sentiment, emotion and toxicity features of the input. Finally, a prediction is made in line 12 and subsequently, the loss is calculated in line 13.

\begin{algorithm}[h!]
    \caption{Sexism Detection}
    \begin{tabular}{ll}
    \textit{Input:} & $t$: A social media post in text \\
    & $y$: The label corresponding to the post $T$ \\
    \textit{Output:} & $\hat{y}$: The predicted label \\
    \end{tabular}
    
    \label{algo:sexism_detection}
    \begin{algorithmic}[1]
    \Function{sexism\_detection}{$T$}
        \State $B \gets RoBERTa\_Tokenizer(T)$
        \State $B = \{b_x\}_{x=1}^{X}$ \Comment{Tokenized input}
        \State $L \gets RoBERTa(B)$ \Comment{Last hidden state representation}
        \State $L = \{e_x\}_{x=1}^{X}$ 
        \State $cls \gets L[:, 0, :]$ \Comment{[CLS] token}
        \State $wla \gets$ word\_level\_attention($cls$)
        \State $cls\_aug \gets$ concatenate($cls$, $wla$)
        
        \State $S \gets Vader(T)$
        \State $E \gets NRXLex(T)$
        \State $T \gets ToxicBert(T)$
        
        \State $Pred \gets$ concatenate($cls\_aug$, $S$, $E$, $T$)
        \State Loss(Pred, $y$)
        \State \Return $Pred$
    \EndFunction
    
    \Function{loss}{$\hat{y}$, $y$}
        \State $ce\_loss \gets CrossEntropy\_Loss(\hat{y}, y)$
        \State $cl\_loss \gets Contrastive\_Loss(cls\_aug, y)$
        \State \Return $ce\_loss + cl\_loss$
    \EndFunction
    \end{algorithmic}
\end{algorithm}

\section{Experimental Evaluations}
In this section, we delineate our experimental setup and present our results, to demonstrate the efficacy of our approach.

\subsection {Experimental Setup}
This section describes the datasets we have utilized, followed by the systems used for contextualizing our performance. Moreover, we also describe the evaluation metrics used to quantify our model's performance. 

\subsubsection{Summarization of Datasets}\label{label:dataset_summary}
\begin{enumerate}[ label=\alph*) ]
\item{
        \textbf{$EXIST 2021$:} The sEXism Identification in Social neTworks (EXIST), 2021 dataset was created by Sanchez et al. \cite{RodrguezSanchez2021OverviewOE} to capture sexism across explicit scenarios and implicit scenarios. The dataset consists of two tasks: Sexism Identification and Sexism Categorization. It contains posts from Twitter and Gab in both English and Spanish. 
        Table-\ref{table:desc_exist_task1} and Table-\ref{table:desc_exist_task2} contain the label distribution across the train and set datasets, for Task-1 and Task-2 respectively.}

\vspace{2mm}
\begin{table}[h!]
\centering
\begin{tabular}{|l|l|l|l|l|}
\hline
\hline
\multicolumn{2}{|c|}{\textbf{Train Set Distribution}} & \multicolumn{2}{c|}{\textbf{Test Set Distribution}} \\ \hline
\textbf{Category} & \textbf{Count} & \textbf{Category} & \textbf{Count} \\ \hline \hline
Non-Sexist & 3,600 & Non-Sexist & 2,087 \\
Sexist & 3,377 & Sexist & 2,281 \\ \hline
\end{tabular}
\caption{Exist 2021 - Dataset Description for Task1}
\label{table:desc_exist_task1}
\end{table}

\vspace{2mm}
\begin{table}[h!]
\centering
\begin{tabular}{|p{2.5cm}|l|p{2.5cm}|l|}
\hline
\multicolumn{2}{|c|}{\textbf{Train Set Distribution}} & \multicolumn{2}{c|}{\textbf{Test Set Distribution}} \\ \hline
\textbf{Category} & \textbf{Count} & \textbf{Category} & \textbf{Count} \\ \hline \hline
non-sexist & 3,600 & non-sexist & 2,087 \\
ideological-inequality & 866 & ideological-inequality & 621 \\
stereotyping-dominance & 809 & stereotyping-dominance & 464 \\
misogyny-non-sexual-violence & 685 & misogyny-non-sexual-violence & 472 \\
sexual-violence & 517 & sexual-violence & 400 \\
objectification & 500 & objectification & 324 \\ \hline
\end{tabular}
\caption{Exist 2021 - Dataset Description for Task2}
\label{table:desc_exist_task2}
\end{table}

To preprocess the dataset, we apply regular expressions to remove the URLs, mentions and hashtags from the posts, as described in Section-3.

    \item{
    \textbf{$MLSC$ }(Multi-label Categorization of Accounts of Sexism using a Neural Framework): This dataset was created by Parikh et al. (2019) \cite{parikh2019multi} as a multi-label classification task. 
The dataset contains 13,023 samples, taken from the \footnote{https://everydaysexism.com } website containing accounts of sexism, with each sample labeled with one or more of the 23 defined sexism labels, mapped to 14 categories.
We split the dataset into 80\% training and 20\% testing dataset, resulting in a training set of 10,418 samples and a testing set of 2,605 samples.
 Table-\ref{table:mlsc_category_distribution} provides a description of the MLSC dataset. }

\begin{table}[h!]
\centering
\renewcommand{\arraystretch}{1.2} 
\setlength{\tabcolsep}{4pt} 
\begin{tabularx}{\columnwidth}{|X |c| c|} 
\hline \hline
\textbf{Category} & \textbf{Train} & \textbf{Test} \\ \hline \hline
Sexual harassment (excluding assault) & 3370 & 871 \\
Attribute stereotyping & 2,184 & 550 \\
Hostile work environment & 2,140 & 520 \\
Role stereotyping & 2,032 & 538 \\
Hyper-sexualization (excluding body shaming) & 1,849 & 471 \\
Other & 1,786 & 454 \\
Sexual assault & 1,526 & 328 \\
Denial or trivialization of sexist misconduct & 1,460 & 371 \\
Moral policing and victim blaming & 1,235 & 332 \\
Internalized sexism & 751 & 203 \\
Body shaming & 418 & 128 \\
Motherhood and menstruation-related discrimination & 422 & 98 \\
Threats & 384 & 116 \\
Slut shaming & 386 & 96 \\ \hline
\end{tabularx}
\caption{MLSC Category-Wise Distribution}
\label{table:mlsc_category_distribution}
\end{table}

\end{enumerate}

\subsection {Experimental Results} \label{label:experiment_results}
To measure the performance of our proposed method, we employ different evaluation metrics, namely accuracy-score, precision, recall, F1-score and macro F1-score, where the macro F1-score denotes the average of the F1-score across all the classes.

\subsubsection{Performance Evaluation on Exist Dataset}
To validate the effectiveness of our approach, we compare it against existing approaches. The comparison results on Exist dataset are shown in Table-\ref{table:exist_task1_results} and Table-\ref{table:exist_task2_results}

\vspace{2mm}
\begin{table}[h!]
\centering
\resizebox{\columnwidth}{!}{%
\begin{tabular}{|l|c|c|c|c|}
\hline
\hline
\textbf{Model} & \textbf{Accuracy} & \textbf{Precision} & \textbf{Recall} & \textbf{Macro F1} \\ \hline \hline
BERT \cite{devlin2018bert} & 0.7301 & 0.7301 & 0.7306 & 0.7300 \\
XLM-R \cite{conneau2019unsupervised} & 0.7587 & 0.7587 & 0.7592 & 0.7586 \\
mBERT \cite{devlin2018bert} & 0.7384 & 0.7384 & 0.7389 & 0.7383 \\
T5 \cite{raffel2020exploring} & 0.6859 & 0.6855 & 0.6858 & 0.6855 \\
LLaMa-3.1 \cite{dubey2024llama} & 0.6604 & 0.6775 & 0.6523 & 0.6445 \\
GPT 3.5 & 0.6772 & 0.4959 & 0.4432 & 0.4291 \\
UPV \cite{RodrguezSanchez2021OverviewOE} & 0.7804 & - & - & 0.7802 \\
BERT\_Ensemble \cite{mazari2024bert} & 0.6775 & 0.6850 & 0.6800 & 0.6750 \\
CLassifiers \cite{radler2023classifiers} & 0.7678 & 0.7674 & 0.7678 & 0.7676 \\
ROH\_NEIL \cite{koonireddy2023roh_neil} & 0.7873 & 0.7870 & 0.7868 & 0.7870 \\
BiLSTM\_CNN \cite{kumar2024hybrid} & 0.6147 & 0.6145 & 0.6147 & 0.6144 \\
\hline
\textbf{ASCEND} & \textbf{0.7984} & \textbf{0.7922} & \textbf{0.7900 }& \textbf{0.7905} \\
\hline
\end{tabular}
}
\caption{Exist Task-1 Performance Comparison}
\label{table:exist_task1_results}
\end{table}

To validate the performance of our proposed method, we carry out a performance comparison with systems implemented in prior work.
The Exist-2021 dataset uses accuracy as the primary metric in the first task and F1-score as the primary metric in the second task. 

In the first task, ASCEND achieves an accuracy of 79.84\%, a precision of 79.22\%, a recall of 79\% and a macro F1-score of 79.05\%, surpassing baseline methods and current state-of-the-art approaches. We outperform the transformer based approaches by employing contrastive learning -  The proposed approach achieves a performance gain of 6.83\%, 6.21\%, 5.94\% and 6.05\% over the BERT model in the accuracy, precision, recall and macro-F1 metrics. We outperform XLM-R by 3.97\%, 3.35\%, 3.08\% and 3.19\%, mBERT by 6\%, 5.38\%, \ 5.11\%, 5.22\% and T5 by 11.25\%, 10.67\%, 10.42\% and 10.50\% in accuracy, precision, recall and macro-F1 respectively. We also outperform modern large-language-models, namely LLaMa-3.1-8B by 13.8\%, 11.47\%, 13.77\%, 14.6\% and GPT 3.5  by 12.12\%, 29.63\%, 34.68\%, 36.14\% with respect to accuracy, precision, recall and macro-F1 score, respectively. Moreover, our system outperforms the current state-of-the-art approaches by 1.1\%, 0.52\% , 0.32\% and 0.35\% 
in terms of precision, recall, accuracy and F1 respectively. Our method also outperforms the challenge leaders by 1.8\% and 1.03\% in terms of accuracy and macro-F1 respectively. 

\vspace{2mm}
\begin{table}[h!]
\centering
\resizebox{\columnwidth}{!}{
\begin{tabular}{|l|c|l|l|l|}
\hline
\hline
\textbf{Model} & \textbf{Accuracy} & \textbf{Precision} & \textbf{Recall} & \textbf{Macro F1} \\ \hline \hline
BERT \cite{devlin2018bert} & 0.4970 & 0.2786 & 0.1981 & 0.1630 \\
XLM-R \cite{conneau2019unsupervised} & 0.5190 & 0.2973 & 0.1693 & 0.1135 \\
mBERT \cite{devlin2018bert} & 0.5147 & 0.3855 & 0.1914 & 0.1525 \\
T5 \cite{raffel2020exploring} & 0.5167 & 0.4304 & 0.3309 & 0.3465 \\
LLaMa-3.1 \cite{dubey2024llama} & 0.4400 & 0.3407 & 0.3120 & 0.3148 \\
GPT 3.5 & 0.3475 & 0.4959 & 0.4432 & 0.4291 \\
UPV \cite{RodrguezSanchez2021OverviewOE} & 0.6577 & - & - & 0.5787 \\
BERT\_Ensemble \cite{mazari2024bert} & 0.4856 & 0.3860 & 0.1770 & 0.1290 \\
CLassifiers \cite{radler2023classifiers} & 0.5526 & 0.5106 & 0.3644 & 0.3999 \\
ROH\_NEIL \cite{koonireddy2023roh_neil} & 0.5831 & 0.5067 & 0.4107 & 0.4210 \\
BiLSTM\_CNN \cite{kumar2024hybrid} & 0.2646 & 0.1917 & 0.2025 & 0.1652 \\
\hline
\textbf{ASCEND}  & \textbf{0.6691} & \textbf{0.5966} & \textbf{0.5847} & \textbf{0.5885} \\
\hline
\end{tabular}}
\caption{Exist Task-2 Performance Comparison}
\label{table:exist_task2_results}
\end{table}

For the second task, ASCEND achieves an accuracy of 66.01\%, a precision of 59.66\%, a recall of 58.47\% and a macro-F1 score of 58.85\%. This surpasses the transformers - BERT by 17.21\%, 31.8\%, 28.66\%, 42.55\%, XLM-R by 15.01\%, 29.93\%, 41.54\%, 47.5\%, mBERT by 15.5\%, 21.11\%, 39.33\%, 43.5\% and T5 by 15.24\%, 16.62\%, 25.38\%, 24.2\% in terms of accuracy, precision, recall and macro F1 respectively. We also surpass large language models such as LLaMa-3.1 by 22.91\%, 25.59\%, 27.27\%, 27.37\% and GPT 3.5 by 32.16\%, 10.07\%, 14.15\% and 15.94\% with respect to the accuracy, precision, recall and macro-F1 scores. Moreover, our approach also outperforms the current state-of-the-art approaches, who are also the challenge leaders, by 1.24\% and 0.98\% in terms of accuracy and F1 respectively . This is due to the use of contrastive learning and word level attention. By weighing terms in the input sequence, word level attention allows the system to focus on the important terms, relevant to the classification task.

\subsubsection{Performance Evaluation on MLSC Dataset}
We compare our approach to the previously mentioned approaches on the MLSC dataset, the results for which are displayed in Table-\ref{table:mlsc_comparison}. The primary metric used for comparison is macro-F1 score.

\begin{table}[h!]
\centering
\begin{tabular}{|l|l|l|l|}
\hline
\hline \textbf{Model} & \textbf{Precision} & \textbf{Recall} & \textbf{Macro F1} \\ \hline \hline
 BERT \cite{devlin2018bert} & 0.3287 & 0.0547 & 0.0880 \\
 XLM-R \cite{conneau2019unsupervised} & 0.3150 & 0.0350 & 0.0589 \\
 mBERT \cite{devlin2018bert} & 0.1428 & 0.0017 & 0.0034 \\
 T5 \cite{raffel2020exploring} & 0.6222 & 0.1848 & 0.2537 \\
LLaMa-3.1 \cite{dubey2024llama}& 0.5804 & 0.5891 & 0.5842 \\
 GPT 3.5 & 0.6237 & 0.6143 & 0.6186 \\
 Heirarchy \cite{parikh2019multi} & - & - & 0.6850 \\
 BERT\_Ensemble \cite{mazari2024bert} & \textbf{0.8990} & 0.5520 & 0.5600 \\
 CLassifiers \cite{radler2023classifiers} & 0.7107 & 0.1591 & 0.2232 \\
 ROH\_NEIL \cite{koonireddy2023roh_neil} & 0.8735 & 0.1903 & 0.2620 \\
 BiLSTM\_CNN \cite{kumar2024hybrid} & 0.7944 & 0.6289 & 0.6658 \\
 \midrule
\textbf{ASCEND}  & 0.6819 & \textbf{0.8523} & \textbf{0.6896} \\
  \hline
\end{tabular}
\caption{MLSC Performance Comparison}
\label{table:mlsc_comparison}
\end{table}

Our method achieves a precision of 68.19\%, a recall of 85.23\% and a macro-F1 score of 68.96\%, surpassing large-language models more than 22 times the size of our model - LLaMa 3.1 by 9.75\% in terms of precision, 26.3\% in terms of recall and 26.81\% in terms of the F1-score. 
We outperform GPT 3.5 Turbo by 5.82\%, 23.8\% and 7.1\% with respect to the precision, recall and macro-F1 score. By employing thresholding, we reduce the occurrence of false positive and negatives, boosting our performance. Moreover, we outperform the most successful approach described in the paper by 0.46\% with respect to the macro-F1 score. Our proposed approach makes predictions based on multiple aspects of the content - Word level attention representation of the sequence, the contrastive learning embedding space and the Emotion, Sentiment and Toxicity Features extracted from the sequence. Combined, these aspects enable a fine-grained analysis of the input sequence, which is an important aspect for a multi-label classification task.

\subsection{Ablation Analysis}
\subsubsection{Ablation on Contrastive Loss and Attention}
The \autoref{table:abl_att} presents the effect of ablation experiments on the performance of the proposed model by excluding two key components: Contrastive Loss (CL Loss) and Word-Level Attention (WLA). The metrics evaluated—Accuracy, Precision, Recall, and Macro F1-score highlight how these components contribute to the overall performance of the model in detecting sexist content.

\begin{table}[h!]
\centering
\resizebox{\columnwidth}{!}{
\begin{tabular}{|l|c|l|l|l|}
\hline
\hline
\textbf{Model} & \textbf{Accuracy} & \textbf{Precision} & \textbf{Recall} & \textbf{Macro F1} \\ \hline \hline
w/o CL Loss & 0.7762 & 0.7743 & 0.7735 & 0.7768 \\
w/o WLA & 0.7793 & 0.7771 & 0.7762 & 0.7783 \\

\midrule
\textbf{ASCEND} & \textbf{0.7984} & \textbf{0.7922} & \textbf{0.7900 }& \textbf{0.7905} \\
\hline
\end{tabular}}
\caption{Ablation on Contrastive Loss and Attention}
\label{table:abl_att}
\end{table}

When CL Loss is removed, the Macro F1-score decreases to 0.7768, and the accuracy drops to 0.7762. This reduction demonstrates the importance of CL Loss in improving the model’s discriminative capability. CL helps the model distinguish between similar yet distinct representations, making it effective in separating sexist content from non-sexist content. Its absence diminishes the model's ability to learn robust and separable feature representations.

Excluding Word-Level Attention results in an F1-score of 0.7783, with a slight improvement over the model without Contrastive Loss. However, the Accuracy remains lower than the proposed method at 0.7793. Word-Level Attention is crucial for the model to focus on the most relevant tokens in the text, allowing it to capture critical sexist cues. Its exclusion reduces the model’s precision and recall, indicating a decreased ability to prioritize important words in decision-making.

\begin{figure}[h]
    \centering
    \includegraphics[width=1.0\columnwidth]{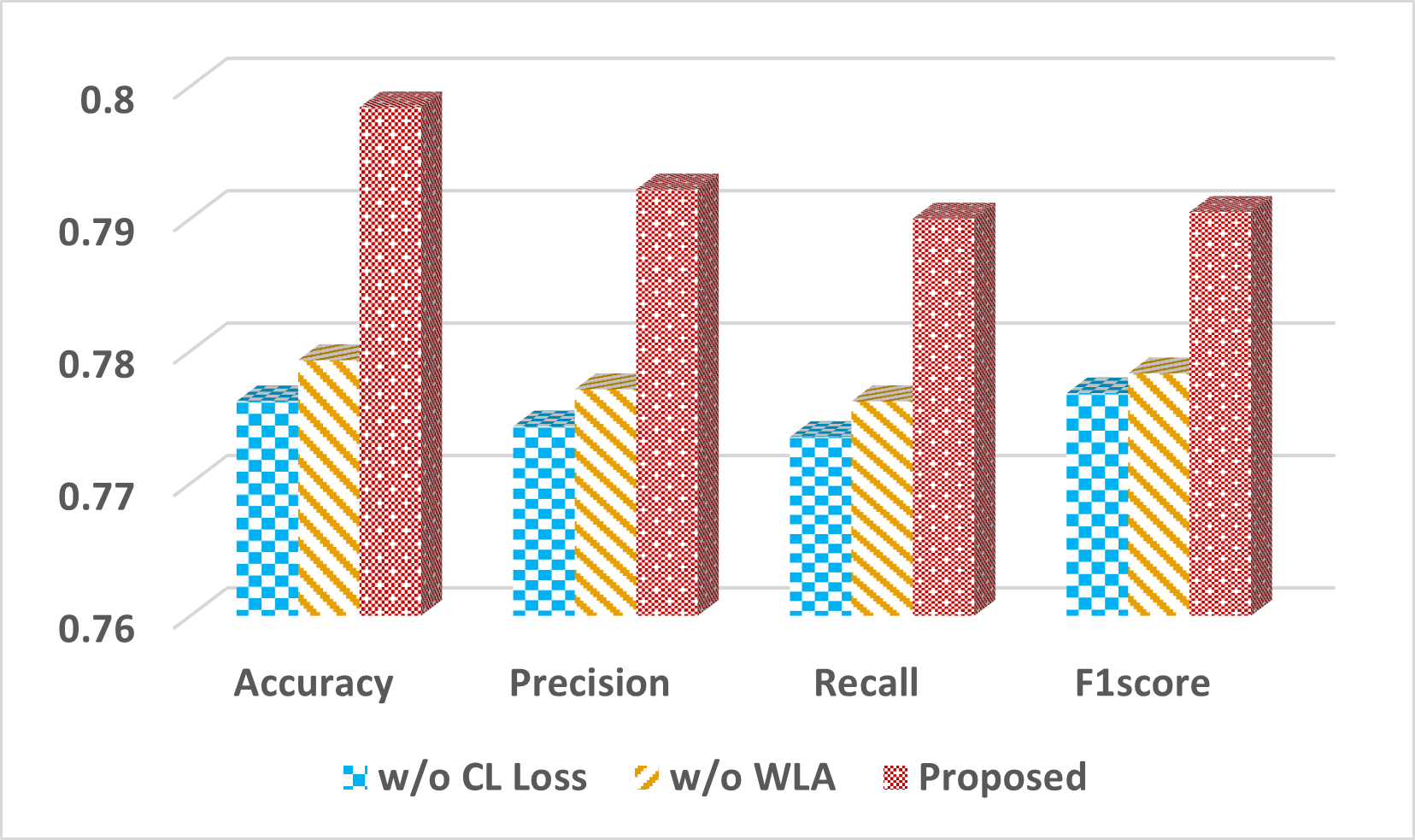}
    \caption{Ablation on Loss and Attention}
    \label{fig:abl_loss}
\end{figure}

The proposed method, which incorporates both Contrastive Loss and Word-Level Attention, achieves the highest performance across all metrics, with an F1-score of 0.7905, Accuracy of 0.7984, Precision of 0.7922, and Recall of 0.79. The combined use of these components enables the model to better represent and focus on sexist content while minimizing irrelevant information, leading to superior detection capability.

In conclusion, the ablation study highlights the complementary roles of Contrastive Loss and Word-Level Attention in enhancing the model's performance. Contrastive Loss aids in learning distinct feature representations, while Word-Level Attention ensures the model focuses on the most critical parts of the input. The integration of both components in the proposed method results in a significant improvement in the detection of sexist language. A graphical representation of the ablation results on loss and attention is shown in \autoref{fig:abl_loss}.

\subsubsection{Ablation on Features}
The \autoref{table:abl_features} demonstrates the influence of different feature sets—Emotion, Sentiment, and Toxicity—on the performance of the proposed model for detecting sexist content. The evaluation metrics, including Accuracy, Precision, Recall, and F1-score, reflect how each feature set contributes to the model's success. The results highlight the complementary role of these features, as their exclusion leads to varying degrees of performance degradation.

\begin{table}[h!]
\centering
\resizebox{\columnwidth}{!}{
\begin{tabular}{|l|c|l|l|l|}
\hline
\hline
\textbf{Model} & \textbf{Accuracy} & \textbf{Precision} & \textbf{Recall} & \textbf{Macro F1} \\ \hline \hline
w/o Emotion features & 0.7912 & 0.7852 & 0.7844 & 0.7836 \\
w/o Sentiment features  & 0.7931 & 0.7824 & 0.7835 & 0.7823 \\
w/o Toxicity features  & 0.7891 & 0.7803 & 0.7784 & 0.7782 \\
\midrule
\textbf{ASCEND} & \textbf{0.7984} & \textbf{0.7922} & \textbf{0.7900 }& \textbf{0.7905} \\
\hline
\end{tabular}}
\caption{Ablation on Features}
\label{table:abl_features}
\end{table}

When emotion features are excluded, the F1-score drops to 0.7836, demonstrating the significance of these features in capturing the nuanced expressions of sexist language. Emotion-related information enables the model to identify subtle linguistic patterns tied to the intensity and context of sexist content, which are harder to capture with other features.

Excluding sentiment features results in an F1-score of 0.7823, indicating that sentiment features also play a crucial role, albeit slightly less impactful than emotion features. Sentiment helps differentiate between positive, negative, and neutral tones in text, which can assist in recognizing sexist content. However, its effect is somewhat less pronounced compared to emotion.

\begin{figure}[h]
    \centering
    \includegraphics[width=1.0\columnwidth]{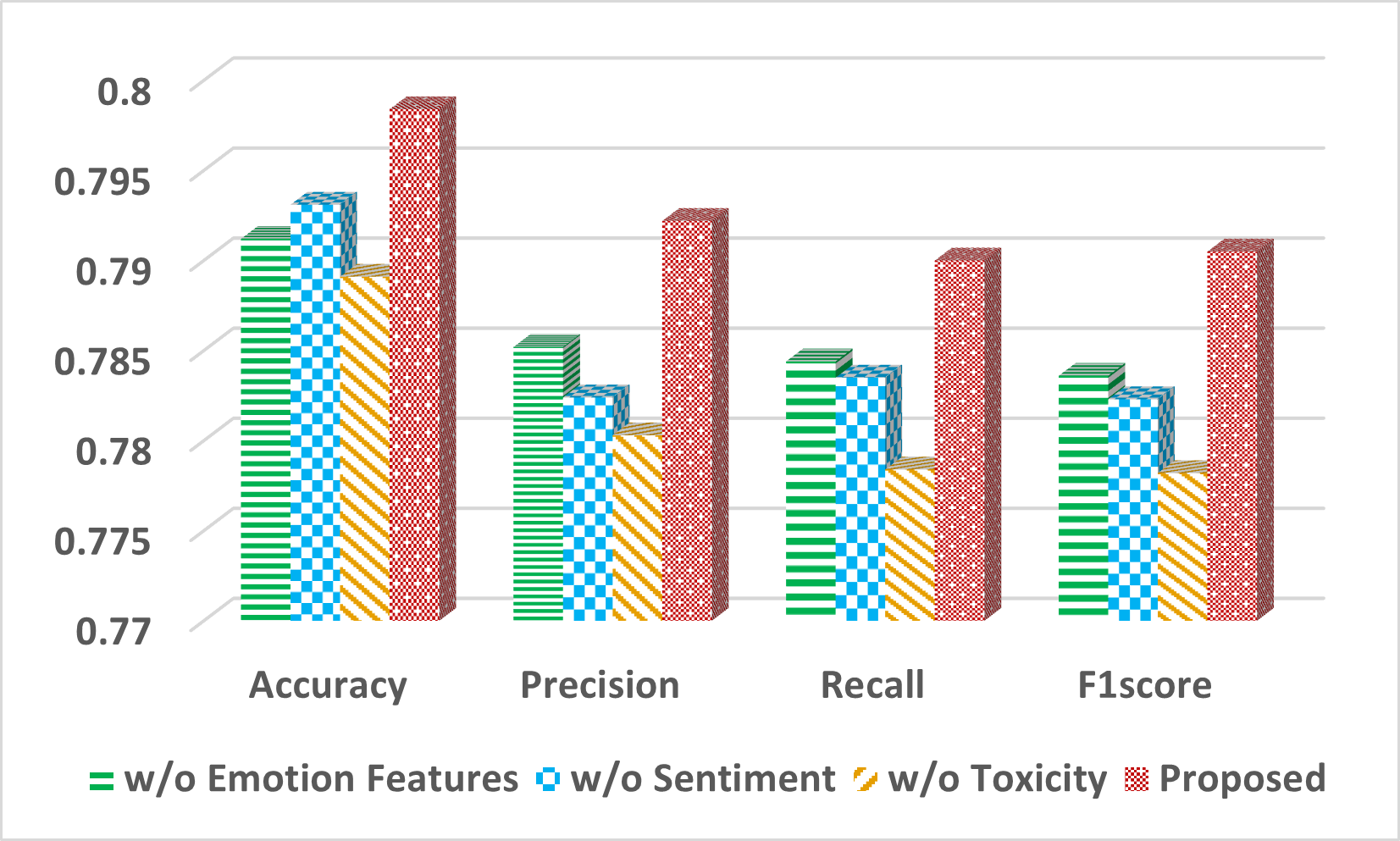}
    \caption{Ablation on Features}
    \label{fig:abl_features}
\end{figure}

The exclusion of toxicity features has the most pronounced negative impact, with the F1-score falling to 0.7782—the lowest among the ablation experiments. Similarly, Accuracy, Precision, and Recall also drop significantly. This finding underscores the critical role of toxicity features, which are directly tailored to identifying sexist patterns, making them indispensable for the model's performance.

The proposed model, which integrates all three feature sets—Emotion, Sentiment, and Toxicity—achieves the best results across all metrics, with an F1-score of 0.7905, Accuracy of 0.7984, Precision of 0.7922, and Recall of 0.79. These results demonstrate the synergistic effect of combining these features, as their interplay enhances the model's ability to detect sexist content comprehensively.

In conclusion, the feature ablation analysis reveals the individual and combined importance of emotion, sentiment, and toxicity features. While each feature set contributes meaningfully to the model's performance, the integration of all three enables the model to achieve superior accuracy and reliability in detecting sexist content. A graphical representation of the ablation results on features is shown in \autoref{fig:abl_loss}.

\subsubsection{Ablation on Threshold}
\begin{figure}
    \centering
    \includegraphics[width=1.0\columnwidth]{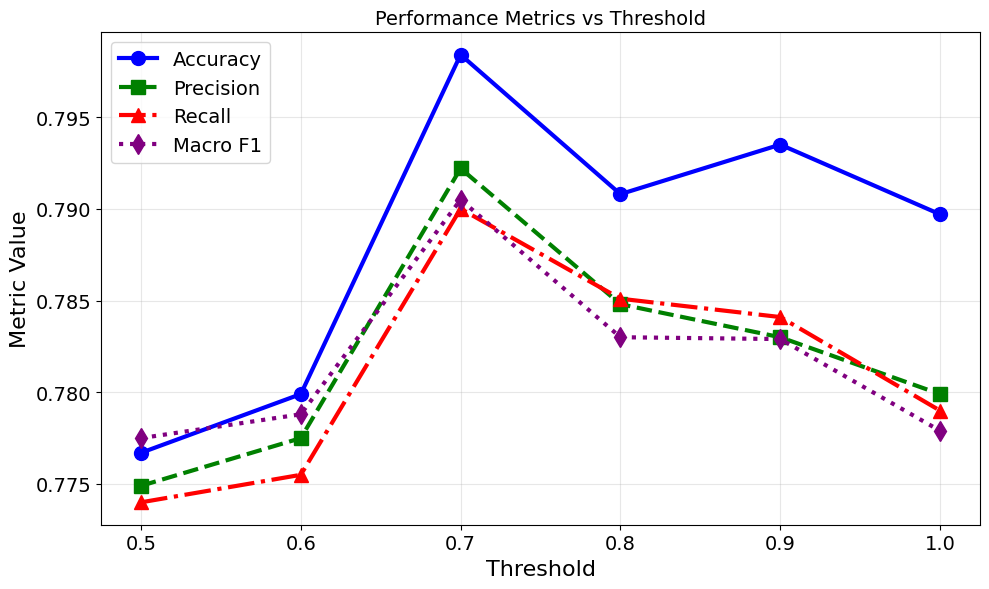}
    \caption{Ablation on Threshold}
    \label{fig:chart}
\end{figure}

The \autoref{fig:chart} presents the impact of varying cosine similarity thresholds on the performance metrics (Accuracy, Precision, Recall, and F1-Score) for sexist content detection. The threshold represents the cosine similarity between sample pairs \((S_i, S_j)\) in a batch, and samples are considered positive matches only if the similarity exceeds the threshold and the corresponding entry in the similarity matrix \([matrix[i][j]]\) is 1. This method ensures that both contextual and semantic similarity between pairs are considered during detection.

\begin{table*}[h]
\footnotesize
\begin{tabularx}{\textwidth}{|p{3.0cm}|X|X|p{2.5cm}|X|X|}
\hline
& Post 1 & Post 2 & Post 3 & Post 4 & Post 5 \\ \hline
Post Details & 
@UNWomen UK @MunroeBer gdorf You're a trans-woman, not A WOMAN. Stop mansplaining women's rights, period. & 
I asked Lana about her new character in "Why Women Kill" Rita Castillo and I told her that I couldn't wait to see the outfits and the hair styles and she said "look at these her nails" \#LanaParilla &
@70SNEWANGEL Can the fellas participate or is this just for the ladies/Non binary people because I don’t wanna get clowned.	 &
Si dice burradas como que Marilyn Monroe was just a dumb blonde. AHÍ NO ES (Translation: if he says stupid things like Marilyn Monroe was just a dumb blonde. THAT'S NOT THERE.) &
@likkeeerr He estado en Inglaterra, y te puedo asegurar que hay gente muy fea con ojos azules. (Translation: @likkeeerr I have been to England, and I can assure you that there are very ugly people with blue eyes.) \\ \hline
Ground Truth & 
Task1:sexist \newline Task2: ideological-in equality &
Task1:non-sexist \newline Task2:\newline non-sexist &
Task1:sexist \newline Task2: ideological-in equality &
Task1:sexist \newline Task2: stereotyping-dominance &
Task1:non-sexist \newline Task2:\newline non-sexist \\ \hline
Model &Prediction &Prediction &Prediction &Prediction &Prediction \\ \hline    
\textit{BERT\_Ensemble \cite{mazari2024bert}} & Task1:\checkmark \newline Task2:\xmark & Task1:\checkmark \newline Task2:\checkmark & Task1:\checkmark \newline Task2:\xmark & Task1:\xmark \newline Task2:\xmark & Task1:\xmark \newline Task2:\xmark \\ \hline
\textit{CLassifiers \cite{radler2023classifiers}} & Task1:\xmark \newline Task2:\xmark & Task1:\checkmark \newline Task2:\checkmark & Task1:\checkmark \newline Task2:\checkmark & Task1:\checkmark \newline Task2:\xmark & Task1:\checkmark \newline Task2:\checkmark \\ \hline
\textit{ROH\_NEIL \cite{koonireddy2023roh_neil}} & Task1:\checkmark \newline Task2:\checkmark & Task1:\checkmark \newline Task2:\checkmark & Task1:\checkmark \newline Task2:\checkmark & Task1:\checkmark \newline Task2:\xmark & Task1:\xmark \newline Task2:\xmark \\ \hline
\textit{BiLSTM\_CNN \cite{kumar2024hybrid}} & Task1:\checkmark \newline Task2:\xmark & Task1:\xmark \newline Task2:\xmark & Task1:\checkmark \newline Task2:\checkmark & Task1:\xmark \newline Task2:\checkmark & Task1:\xmark \newline Task2:\xmark \\ \hline
\textit{LLaMa-3.1 \cite{dubey2024llama}} & Task1:\checkmark \newline Task2:\checkmark & Task1:\xmark \newline Task2:\xmark & Task1:\checkmark \newline Task2:\checkmark & Task1:\checkmark \newline Task2:\checkmark & Task1:\checkmark \newline Task2:\xmark \\ \hline

\textit{GPT 3.5 } & Task1:\checkmark \newline Task2:\checkmark & Task1:\checkmark \newline Task2:\checkmark & Task1:\checkmark \newline Task2:\checkmark & Task1:\xmark \newline Task2:\xmark & Task1:\checkmark \newline Task2:\checkmark \\ \hline

\textit{\textbf{ASCEND}} & Task1:\checkmark \newline Task2:\checkmark & Task1:\checkmark \newline Task2:\checkmark & Task1:\checkmark \newline Task2:\checkmark & Task1:\checkmark \newline Task2:\checkmark & Task1:\checkmark \newline Task2:\checkmark \\ \hline
\end{tabularx}
\caption{Qualitative Analysis}
\label{table:qualitative_analysis}
\end{table*}

At a low threshold of 0.5, the model achieves moderate performance, with an F1-score of 0.7775. This indicates that many sample pairs are classified as positives due to the low similarity requirement, leading to relatively balanced Precision (0.7749) and Recall (0.774).

As the threshold increases to 0.6, the performance improves slightly. The F1-score increases to 0.7788, indicating that higher cosine similarity requirements enhance the model's ability to identify more contextually aligned pairs without significantly sacrificing either Precision (0.7775) or Recall (0.7755).

At a threshold of 0.7, the model achieves its highest performance, with an Accuracy of 0.7984 and an F1-score of 0.7905. This threshold strikes the optimal balance between Precision (0.7922) and Recall (0.79), suggesting that pairs classified as positives at this level share a strong semantic similarity, effectively improving sexist content detection.

However, further increasing the threshold to 0.8 results in a decline in performance metrics. The F1-score decreases to 0.783, with Precision dropping to 0.7848 and Recall to 0.7851. This trend indicates that the stricter similarity requirement excludes some true positive pairs, reducing the model's ability to generalize effectively.

At thresholds of 0.9 and 1, the metrics continue to decline slightly. At a threshold of 1, the F1-score falls to 0.7779, with Precision at 0.7799 and Recall at 0.779. These results suggest that the overly restrictive similarity criteria lead to a higher number of false negatives, as only nearly identical pairs are classified as positives.

Overall, the results highlight that a cosine similarity threshold of 0.7 provides the best trade-off between Precision and Recall, making it the most effective setting for detecting sexist content based on pairwise sample similarity. This threshold ensures a balance between identifying meaningful semantic relationships while minimizing the risk of misclassification.

\subsection{Qualitative Analysis}

Table \ref{table:qualitative_analysis} presents a qualitative comparison of various models on five sample posts from the dataset, evaluating their performance on Task 1 (sexist vs. non-sexist classification) and Task 2 (sexism subcategory classification). The results highlight key differences in model capabilities, particularly in handling subtle hateful content.

Overall, ASCEND demonstrates the highest accuracy, correctly classifying both tasks across all examples. In contrast, models such as GPT-3.5 and LLaMa-3.1 perform well but exhibit occasional misclassifications, particularly in Task 2. Traditional models, including BERT\_Ensemble and BiLSTM\_CNN, struggle significantly with sexism subcategory identification, misclassifying multiple instances. Task 2 proves more challenging than Task 1, as evident in cases like Post 1, where many models correctly identify it as sexist but fail to classify it under ideological inequality. Similarly, Post 4, which involves cultural stereotyping, is misclassified by most models except ASCEND.

Several failure cases are also observed. Post 2, a non-sexist statement, is correctly classified by most models, but BiLSTM\_CNN incorrectly labels it as sexist, indicating potential overfitting to certain linguistic patterns. Post 5, another non-sexist post, is misclassified by ROH\_NEIL and BERT\_Ensemble, highlighting their difficulty in distinguishing neutral statements from actual hate speech. These observations suggest that while transformer-based models are generally more effective than traditional models, their performance varies based on the complexity of the sexism subcategories.

ASCEND's superior performance may be attributed to its ability to leverage multi-modal cues or a deeper contextual understanding, allowing it to capture nuanced patterns in sexist language. Its robustness in correctly classifying both Task 1 and Task 2 labels across all examples makes it a promising approach for real-world hateful content moderation. These results emphasize the importance of fine-grained classification in hate speech detection. While large-scale models like GPT-3.5 and LLaMa-3.1 perform well, ASCEND consistently outperforms all other methods, particularly in identifying sexism subcategories. Future work could explore the role of explainability in these models to further enhance interpretability and fairness.

\vspace{0.1in}
\section{Conclusion}
\label{sec:sd_cnfw}
In this paper, we propose an adaptive supervised contrastive learning framework to identify sexist content in an implicit environment. Our method employs thresholding to reduce the occurrences of false positives and false negatives. Furthermore, we enhance vector representations in the embedding space by using the word-level-attention vector representation of a textual sequence. We also extract the emotion, sentiment, and toxicity features of input during the classification task. Together, these modules are capable of detecting, classifying, and categorizing the text into a fine-grained taxonomy used in previous works. We demonstrate that our system outperforms the baseline and current state-of-the-art methods across three tasks, distributed among two datasets. Our model achieves better results on all evaluation metrics, as shown by our experimental results.




\bibliographystyle{ieeetr}
\bibliography{sn-bibliography}

\end{document}